%% file: Optimal_Minibatch.tex
\def\year{2020}\relax
\documentclass[letterpaper]{article} %DO NOT CHANGE THIS
\usepackage{aaai20}  %Required
\usepackage{times}  %Required
\usepackage{helvet}  %Required
\usepackage{courier}  %Required
\usepackage{url}  %Required
\usepackage{graphicx}  %Required
\usepackage{booktabs}
\usepackage{amsmath}        % PERRONE added for cases in equation environments
\usepackage{color, colortbl}% PERRONE added for color in table environments

\newcommand{\NU}{N_{\rm{Update}}}
\newcommand{\TU}{T_{\rm{Update}}}
\newcommand{\TC}{T_{\rm{C}}}
\definecolor{Gray}{gray}{0.9}

\title{Optimal Mini-Batch Size Selection for Fast Gradient Descent}
\author{
Michael P. Perrone \\
IBM T.J. Watson Research Center\\
Yorktown Heights, NY 10598\\
\texttt{mpp@us.ibm.com}\\
\And
Haidar Khan\\
Amazon Alexa\\
New York, NY 10001\\
\texttt{khhaida@amazon.com}
\And
Changhoan Kim \\
Morgan Stanley \\
New York, NY 10001\\
\AND
Anastasios Kyrillidis \\
CS Dept. 
Rice University\\
Houston, TX 77005\\
\texttt{anastasios@rice.edu}
\And
Jerry Quinn \\
IBM T.J. Watson Research Center\\
Yorktown Heights, NY 10598\\
\And
Valentina Salapura \\
IBM T.J. Watson Research Center\\
Yorktown Heights, NY 10598\\
}

\begin{document}

\maketitle

\begin{abstract}
This paper presents a methodology for selecting the mini-batch size that minimizes Stochastic Gradient Descent (SGD) learning time for single and multiple learner problems. By decoupling algorithmic analysis issues from hardware and software implementation details, we reveal a robust empirical inverse law between mini-batch size and the average number of SGD updates required to converge to a specified error threshold.
Combining this empirical inverse law with measured system performance, we create an accurate, closed-form model of average training time and show how this model can be used to identify quantifiable implications for both algorithmic and hardware aspects of machine learning.
%Using this model, we explain that minimizing the time to compute an epoch, or any fixed number of updates, does not necessarily minimize the total training time because it neglects the dependence of convergence time on mini-batch size. 
We demonstrate the inverse law empirically, on both image recognition (MNIST,
CIFAR10 and CIFAR100) and machine translation (Europarl) tasks, and provide a theoretic
justification via proving a novel bound on mini-batch SGD training. 
% by providing specific guidance
% (1) to system designers on how to best allocate limited system
% resources for optimal SGD convergence time; and (2) to learning
% algorithm designers on which global algorithmic parameters drive optimal
% SGD convergence time.
% Using this
% relationship, we define an optimal data-parallel scaling method which
% outperforms both strong scaling and the commonly used weak scaling seen
% in much of machine learning scaling literature. 
\end{abstract}

%%%%%%%%%%%%%%%%%%%%%%%%%
\section{Introduction}

% At its best, the study of machine learning is a principled scientific exploration of the properties of learning, coupled with solid engineering to take these ideas from the laboratory to real-world practice. 
% However, machine learning can sometimes appear to be voodoo 
% %, or a kind New Age alchemy chasing after the philosopher's stone to turn base data into consciousness 
% \cite{recht2017reflections}: \emph{e.g.}, hyperparameter tuning is often more of an art than a science, with significant time and effort spent searching for the best parameters, often without sufficient solid evidence nor theoretical underpinning to guide the search.  

In this paper, we present an empirical law, with theoretical justification, linking the number of learning iterations to the mini-batch size.  From this result, we derive a principled methodology for selecting mini-batch size w.r.t. \emph{training performance}\footnote{Any connections with testing/generalization performance are left for future work.} for data-parallel machine learning. 
This methodology saves training time and provides both intuition and a principled approach for optimizing machine learning algorithms and machine learning hardware system design.
Further, we use our methodology to show that focusing on weak scaling can lead to suboptimal training times because, by neglecting the dependence of
convergence time on the size of the mini-batch used, weak scaling does not always minimize the training time. 
All of these results derive from a novel insight presented in this paper, that
understanding the average algorithmic behavior of learning, decoupled
from hardware implementation details, can lead to deep insights into machine learning.

%%%%%%%%%%%%%%%%%%%%%%%%
%Furthermore, the ultimate success of Stochastic Gradient Descent (SGD) machine
%learning for truly large, real-world learning problems depends on the
%ability to efficiently explore a vast space of algorithmic and model
%topology choices to build useful systems. The assessment of each choice
%in turn can require optimization in billion-dimensional parameter
%spaces. Thus, designing efficient hardware to run these learning
%problems is critical.

Our results have direct relevance to on-going research that accelerate training time.
For example, significant research effort has been focused on
accelerating mini-batch SGD \cite{dekel2012optimal,keskar2016large,li2014efficient}, primarily focused on
faster hardware \cite{coates2013deep,krizhevsky2014one,tan2011fast,chetlur2014cudnn}, parallelization using multiple learners
\cite{cho2017powerai,goyal2017accurate,dean2012large}, and improved algorithms and system designs for
efficient communication (\emph{e.g.}, parameter servers, efficient passing of
update vectors \cite{goyal2017accurate,watcharapichat2016ako,lian2015asynchronous,recht2011hogwild,seide2014parallelizability,zhang2015deep,li2014scaling}, \emph{etc.} 
To assess the impact of these acceleration methods, published research typically evaluates
parallel improvements based on the time to complete an epoch for a fixed
mini-batch size, what is commonly known as ``weak'' scaling \cite{cho2017powerai,goyal2017accurate,watcharapichat2016ako,recht2011hogwild}.

The next section explains how the insight of decoupling algorithmic and implementation details is used to develop a close-form model for learning convergence time.  We then derive implications for hardware system design. These implications are followed by sections with experimental support and theoretical justification for the empirical law connecting learning iterations with mini-batch size.  We close with a discussion section.

%In Section~\ref{sec:Decompose}, we
%make a key observation about optimizing SGD convergence time and
%describe how it enables the analysis performed in this paper. In Section~\ref{sec:ModelingT},
%we derive the average convergence time by modeling both the mini-batch update time 
%and the number of SGD updates required to converge.
%In Section~\ref{sec:Optimalmini-batch}, we show how to select the optimal mini-batch size for single and multi-learner cases.
%% In Section~\ref{sec:Scaling}, we derive optimal scaling for data-parallel SGD learning. 
%% In Section~\ref{sec:SystemDesign}, we explain how optimal scaling can be used to improve
%% system design, and to guide the creation of novel algorithms that benefit parallel
%% learning.
%In Section~\ref{sec:Results}, we
%provide experimental results that demonstrate the robustness of the
%inverse law linking mini-batch size and the average number of
%SGD updates required to converge. In Section~\ref{sec:Theory},
%we support our empirical law by deriving a theoretical
%bound for mini-batch SGD learning that exhibits the same inverse
%relationship. And in Section~\ref{sec:Summary}, we summarize and discuss
%additional implications for future research.
%
%%%%%%%%%%%%%%%%%%%%%%%%%
\section{Modeling SGD Convergence Time\label{sec:Decompose}}

Given a learning problem represented by a data set, SGD as the learning
algorithm, and a learning model topology, we define the learning time,
$\TC$, to be the average total time required for SGD to converge to a
specified achievable training error threshold. 
The average is over all possible sources of randomness in the
process, including random initializations of the model, ``noise'' from the SGD
updates, noise in the system hardware, etc. 
Focusing on the average
learning behavior allows us to identify fundamental properties of the
learning process. In particular, we can write the \emph{analytical complexity} as the product of \emph{iteration complexity}, $\NU$, and the \emph{per iteration computational complexity}, $\TU$.
\begin{equation}
\label{eqn:Decompose}
\TC=\NU \cdot \TU
\end{equation}
In other words, $\NU$ is the average number of updates
required to converge, and $\TU$ is the average
time to compute and communicate one update. 
%Note that $\NU$ is independent of
%implementation details, such as the choice of hardware, the choice of software, and the
%speed with which SGD updates are calculated.\footnote{Here we require that the model update is the same regardless of the hardware, system configuration, software implementation, etc.  By way of analogy, $2+2$ equals 4 regardless of how the calculation is implemented in hardware.  And, specifically, $\NU$ is independent of $\TU$.  Another way of saying this is that although $\TU$ depends on the algorithm, the algorithm does not depend on $\TU$.}
%$\NU$ depends only on the data, the learning
%algorithm used and the learning model topology. Interestingly, $\NU$ can be thought of as
%the difficulty to learn a task, while $\TU$
%is the difficulty to compute an update.
%%%%%%%%%%%%%%%%%%%%%%%%
%%On the other hand, $\TU$ depends on the choice of computational
%%hardware, and the amount and type of computation required for a single
%%update, \emph{i.e.}, the amount of data used to calculate each update, the
%%model topology, the software implementation of the learning algorithm,
%%and time to communicate SGD updates between the parallel learners of the
%%system. Thus, $\NU$ is independent of all
%%hardware considerations, and for fixed algorithm and model topology,
%%$\TU$ depends only on hardware choices. 

This decomposition of $\TC$ into an implementation-dependent and implementation-independent components is useful because it helps decouple the tasks of understanding how
implementation and algorithmic choices impact learning time, and allows us to
understand algorithmic choices, independent of system design choices.

%%%%%%%%%%%%%%%%%%%%%%%%%
\subsection{Modeling Average Convergence Time\label{sec:ModelingT}}

To analyze SGD convergence, we model $\NU$ as a function of the
mini-batch size, $M$, and model $\TU$ as a function of $M$ and the number of parallel
learners, $P$. 
 All other hyperparameters are held constant. 
% We develop these models below and then use them to derive an optimal mini-batch size.

For simplicity, we refer to ``learner'' parallelism when multiple learners share the task of learning a single model.  
Each learner is mapped to  a compute element from a suitable level of parallelism, \emph{e.g.}, a server, a CPU, a GPU, \emph{etc. }

In general, the level of parallelism
selected will have implications for the software
implementation, communication requirements, and system performance.
However, our analysis below is independent of these details.

%%%%%%%%%%%%%%%%%%%%%%%%%
\subsection{Modeling $\NU(M)$}

Since $\NU$ is independent of the hardware, it
is independent of the number of compute elements used, and therefore
depends only on the mini-batch size, $M$. %\footnote{We hold the data set, learning algorithm and model topology fixed during training.} 
Even with this simplification, measuring $\NU$ from hardware is generally
impractical due to the computational expense of running SGD to
convergence for all values of $M$. Fortunately, there is an easier
way: We have discovered a robust empirical inverse relationship between
$\NU$ and $M$ given by
\begin{equation}
\label{eqn:Inverse}
\NU = N_{\infty} + \frac{\alpha}{M}
\end{equation}
where $N_{\infty}$ and $\alpha$ are empirical parameters depending
on the data, model topology and learning algorithm used. 
%The experimental evidence
%supporting this inverse relationship is present in Section~\ref{sec:Results}, and a
%theoretical justification is presented in Section~\ref{sec:Theory}.

This inverse relationship captures the intuitively obvious result that
even if we compute exact gradients, \emph{i.e.}, even when $M$ equals all
of the data in a given data set, gradient descent still requires a
non-zero number of steps to converge. 
% This fact has very important
% implications for parallelization of SGD algorithms, because it implies
% that there are diminishing returns from increased parallelism, as we
% will see in our analysis below.

Furthermore, the Central Limit Theorem tells us that the variance of the
SGD gradient is inversely proportional to $M$, for large $M$.
Thus, $\NU$ increases approximately linearly
with the SGD gradient variance, and $\alpha$ can be thought of the
system's sensitivity to noise in the gradient.  We define $\alpha$ to be the ``noise sensitivity'' of the algorithm.

%%%%%%%%%%%%%%%%%%%%%%%%%
\subsection{Novelty of this Result}

Numerous papers have addressed the behavior and benefits of mini-batch training \cite{cotter2011better,jain2017parallelizing,bottou2018optimization}.  
These papers frequently derive training error bounds in terms of the number of iterations, and/or the mini-batch size, often exhibiting $1/M$ and $1/N$ related behavior. 
Although these results superficially look similar to Eqn.~\ref{eqn:Inverse}, they
do not address the functional relationship between the number of iterations and $M$.  
Our work is the first to demonstrate this relationship empirically, on a wide variety of learning problems, in a much simpler way.
Furthermore, for some of the results in \cite{cotter2011better,jain2017parallelizing,bottou2018optimization}, it is possible to go beyond what the original authors intended by inverting their results to find relationships between number of iterations and $M$; however, without empirical evidence, one does not know how tight the bounds are in practice, and even if one assumes they are tight, the resulting inversion leads to relationships which are not the same as Eqn.~\ref{eqn:Inverse}.

To make this explicit with an example, consider the second equation on P.~31 of \cite{bottou2018optimization} which, with simplified notation and constants $A$ and $B$,
can be written as
$$E_N \leq A/M + (1-B/M)^{N-1}(E_1-A/M)$$
If we replace $E_N$ with a target error threshold, $\epsilon$, this equation implies the following inequality:
$$N\geq 1+\frac{M}{M-B}\ln\left(\frac{\epsilon M-A}{E_1M-A}\right)$$
which is clearly different from Eqn.~\ref{eqn:Inverse}.
%This relationship is different than the results presented in our paper. 
One can perform similar analyses to the other published results to demonstrate that they are not equivalent Eqn.~\ref{eqn:Inverse}.

%%%%%%%%%%%%%%%%%%%%%%%%%
\subsection{Modeling $\mathbf{\TU(M,P)}$}

Measuring $\TU$ is comparatively
straightforward: One need only run enough iterations of SGD for a single learner to estimate the average
time to perform an update for a specified mini-batch size. This process
is possible because $\TU(M,P)$ is approximately constant throughout SGD learning.
% ; so it need only be
% measured once for each $(M,P)$ pair of interest, which is
% generally a small computation compared to running full SGD learning to convergence.
This approach can be used to compare differences between specific types of
hardware, software implementations, \emph{etc.} 
One then use the measured
$\TU$ to fit an analytic model, along with $\NU$, to model $\TC(M,P)$.

To analyze the generic behavior, we model
\begin{equation}
\TU(M,P) = \Gamma(M) + \Delta(P)
\end{equation}
where $\Gamma(M)$ is the average time to compute an SGD update using
$M$ samples, and $\Delta(P)$ is the average time to communicate
gradient updates between $P$ learners.\footnote{Without loss of generality, we subsume any communication time internal to a single learner (\emph{e.g.}, the time to read/write data from memory) into the computation time.}
If some of the communication between learners
can occur during computation, then $\Delta(P)$ represents the
portion of communication that is not overlapping with computation.\footnote{An efficient SGD system will 
attempt to overlap computation and
communication. For example, in backpropagation, gradient updates for all but the
input layer can in principle be transferred during the calculation of updates for
subsequent layers. In such systems, the communication time,
$\Delta(P)$, is understood to mean the portion that does
not overlap with computation time.}
Since computation and communication are handled by separate hardware, it
is a good approximation to assume that they can be decoupled in this
way.

Since $\Gamma(M)$ typically performs the same amount of computation
for each data sample, one might expect a linear relationship,
$\Gamma(M) = \gamma M$, for some constant,
$\gamma$. However, in
practice, hardware and software implementation inefficiencies lead to a
point where reducing $M$ does not reduce compute time linearly.\footnote{Here we are neglecting the generally insignificant time
required to sum over $M$ data samples.}
This effect can be approximated using
\begin{equation}
\label{eqn:HW}
\Gamma(M) = \gamma\max(M,M_T)
\end{equation}
where $M_T$ is the threshold at which the linear relationship
begins, \emph{i.e.}, the knee in the curve. For example, $M_T$ could be the number of cores per CPU, if
each sample is processed by a different core; or $M_T$ could be 1 if
a single core processes all samples. Ideally, efficient SGD hardware
systems should achieve low $\gamma$ and $M_T$. In practice, an
empirical measurement of this relationship provides more fidelity; but
for the purposes of this paper, this model is sufficient.

For $P = 1$, the communication time is zero, \emph{i.e.}, $\Delta(P)=0$. 
For $P > 1$, $\Delta(P)$ depends on various
hardware and software implementation factors. We assume model updates exploit an optimized communication protocol, 
such as the Message Passing Interface
(MPI) function \texttt{MPIAllReduce()} \cite{kumar2016optimization} on a high-end compute cluster.
Such systems provide a powerful network switch and an efficient
\texttt{MPIAllReduce()} implementation that delivers near perfect scaling of
\texttt{MPIAllreduce()} bandwidth, and so communication time is approximately constant, \emph{i.e.},
$\Delta(P)=\delta$ for some constant $\delta$ approximately inversely proportional to
the aggregate bandwidth between learners. 
For comparison purposes, a synchronous parameter server
has a communication time that grows linearly with $P$, \emph{i.e.}, $\Delta(P) = \delta P$.

%%%%%%%%%%%%%%%%%%%%%%%%%
\subsection{Modeling $\mathbf{\TC(M,P)}$}

Using Eqn.~\ref{eqn:Decompose} to combine our estimates for $\NU$, 
$\TU$ yields the following general
approximation to the total convergence time for SGD running on $P$
parallel learners:
%\begin{small}
\begin{equation}
\label{eqn:TC}
\TC(M,P) = \left(N_{\infty} + \frac{\alpha}{M}\right)\left[\gamma\max\left( \frac{M}{P},M_T \right) + \delta \right]
\end{equation}
%\end{small}
We can now use this model to optimize training time and analyze the impact of system design on training time
in numerous ways.  
\emph{E.g.}, in the experiments we show how to 
select the mini-batch size that minimizes SGD training time.
% In Section~\ref{sec:Scaling}, we extend this result to multi-learner 
% data-parallel machine learning to understand the scaling behavior of SGD.

Note that Eqn.~\ref{eqn:TC} relies on certain assumptions about the hardware
that might not be true in general, \emph{e.g.}, that $\delta$ is a constant.
We have chosen these assumptions to simplify the analysis; but in
practice, one can easily choose a different model for $\TC$, or even measure the
exact form of $\TC$, and still follow through with the analysis below.

One final consideration arises regarding cross-validation (CV) since SGD
training is rarely performed without some form of CV stopping criterion.
We can accommodate the effect of CV in our model by including a CV term,
such that
%\begin{small}
\begin{equation}
\Gamma(M) = \gamma N\max(M,M_T) + \gamma_{\rm CV}\max(M_{\rm CV},M_T)
\end{equation}
%\end{small}
where $N$ is the number of SGD updates per CV calculation and
$M_\text{CV}$ is the number of CV samples to calculate. For
simplicity, we ignore CV in our analysis below, but the analysis follows
the same path as before. Additionally, the calculation of a CV subset
adds virtually no communication, since the parallel learners computing the
CV estimate communicate only a single number when they are done.

%%%%%%%%%%%%%%%%%%%%%%%%%
\section{Optimal Mini-Batch Size Selection\label{sec:Optimalmini-batch}}

For the single learner case ($P=1$), there is no inter-learner communication cost, so Eqn.~\ref{eqn:TC} yields 
\begin{equation}
\TC(M,1) = \left(N_{\infty} + \frac{\alpha}{M}\right)\gamma\max\left(M,M_T \right).
\end{equation}
%We now have a closed-form approximation to the average convergence time on a single learner as a function of the mini-batch size.  
Optimizing over $M$, we find the optimal mini-batch size to be
\begin{equation}
M_{\rm Opt} = M_T.
\end{equation}
One can easily identify $M_T$ for a given system by timing a few SGD updates per value of $M$ to obtain an estimate of $\TU(M)$, and then selecting the knee in the $\TU(M)$ curve as an estimate for $M_T$.  If the simple model used in Eqn.~\ref{eqn:HW} is not accurate for a given system configuration, the methodology below can be used.

\begingroup
\setlength{\tabcolsep}{2pt}
\begin{table}[ht]
\centering
%\caption{Method}
%\label{tab:Methodology}
\begin{tabular}{ll}
 \toprule
 \multicolumn{2}{l}{\textbf{Methodology:} Optimal Mini-Batch Size Selection}\\
 \midrule
   1. & For a range of $M$ values: \\ & \quad Measure $\TU(M)$ over a few SGD updates. \\
   2. & For a least two values of $M$: \\ & \quad Estimate $\NU(M)$ by running to convergence. \\
   3. & Fit $N_\infty$ and $\alpha$ to estimate $\NU(M)$ values.\\
   4. & Use $\NU(M,N_\infty,\alpha)$, $\TU(M)$ to select $M_{\rm Otp}$.\\
 \bottomrule
\end{tabular}
\end{table}
\endgroup

%We recommend that machine learning practitioners assess their system configure in this way before selecting %hyperparameters, like mini-batch size.

For the multiple learning case ($P\geq 1$), the optimal $M$ is given by
\begin{equation}
\label{eqn:MOpt}
M_{\rm Opt}(P) = \max\left( \sqrt{\frac{\alpha\delta P}{N_\infty\gamma}}~~,~~M_T P\right)
\end{equation}
which demonstrates that linearly increasing the training data with the number of learners (\emph{i.e.}, ``weak scaling'') is not always the optimal choice because $M_TP$ can be less than $\sqrt{\alpha\delta P/N_\infty\gamma}$.  Note that the methodology above can also be used to optimize $M$ in the multi-learner case.

% \begin{remark}
% Numerous papers have addressed the behavior and benefits of mini-batch training \cite{cotter2011better,jain2017parallelizing,bottou2018optimization}.  
% These papers frequently derive training error bounds in terms of the number of iterations, and/or the mini-batch size, often exhibiting $1/M$ and $1/N$ related behavior. 
% However these results do not address the functional relationship between the number of iterations and $M$.  
% Our work is the first to demonstrate this relationship empirically, on a wide variety of learning problems, in a much simpler way.
% Furthermore, for some of the results in \cite{cotter2011better,jain2017parallelizing,bottou2018optimization}, it is possible to go beyond what the original authors intended by inverting their results to find relationships between number of iterations and $M$; however, without empirical evidence, one does not know how tight the bounds are in practice, and even if one assumes they are tight, the resulting inversion leads to relationships which are not the same as the $\alpha/M + N_\infty$ relationship presented here.
% \end{remark}

%%%%%%%%%%%%%%%%%%%%%%%%%
%%%%%%%%%%%%%%%%%%%%%%%%%
%%%%%%%%%%%%%%%%%%%%%%%%%
\input{Sections/DataParallelScaling}

%%%%%%%%%%%%%%%%%%%%%%%%%
%%%%%%%%%%%%%%%%%%%%%%%%%
%%%%%%%%%%%%%%%%%%%%%%%%%
\input{Sections/SystemDesign}

%%%%%%%%%%%%%%%%%%%%%%%%%
\section{Experimental Results}\label{sec:Results}
We have observed that, to a reasonable approximation, the relationship
\begin{equation}
\NU = N_{\infty} + \frac{\alpha}{M}
\end{equation}
persists over a broad range of $M$, and a variety of machine learning
dimensions, including the choice of learning domain (image recognition and machine translation),
data set, model topology, number of
classes, convergence threshold and learning rate. This section describes
the methodology used to assess this relationship and the results obtained.

\begin{table*}[!ht]
\centering
\caption{Image recognition rraining experiments performed.}
\label{tab:Experiments}
\begin{tabular}{llllll}
 \rowcolor{Gray}
 \toprule
 \textbf{Data Set} & \textbf{Model} & \textbf{\# Parameters} & \textbf{\#
Layers} & \textbf{Learning Rate} & \textbf{$M$ }\tabularnewline
 \midrule
% \endhead
 MNIST & Small -- LeNet \cite{lecun1995learning} & 43,158 & 4 & 0.01 & 1 - 1024\tabularnewline
 & Medium -- LeNet \cite{lecun1995learning} & 169,506 & 4 & 0.01, 0.05 & 1 - 1024\tabularnewline
 & Even/Odd -- LeNet \cite{lecun1995learning} & 169,506 & 4 & 0.05 & 1 - 1024\tabularnewline
 & Large -- LeNet \cite{lecun1995learning} & 671,802 & 4 & 0.01 & 1 - 1024\tabularnewline
 \midrule
 CIFAR10 & Small -- LeNet \cite{lecun1995learning} & 487,458 & 4 & 0.05 & 1 - 1024\tabularnewline
 & Medium -- VGG \cite{simonyan2014very}& 1,125,090 & 9 & 0.01, 0.05, Adam & 1 -
 1024\tabularnewline
 & Large -- VGG \cite{simonyan2014very}& 1,862,754 & 11 & 0.025 & 1 - 1024\tabularnewline
 & ResNet \cite{he2016deep}& 270,410 & 20 & 0.05 & 1 - 1024\tabularnewline
 \midrule
 CIFAR100 & Small -- LeNet \cite{lecun1995learning} & 487,458 & 4 & 0.05 & 16 -
 1024\tabularnewline
 & Medium -- VGG \cite{simonyan2014very}& 1,125,090 & 9 & 0.05 & 16 - 1024\tabularnewline
 & Large - VGG \cite{simonyan2014very}& 1,862,754 & 11 & 0.025, 0.05 & 16 - 1024\tabularnewline
 \bottomrule
\end{tabular}
\end{table*}

\subsection{Image Recognition Task}
To measure the robustness of our observations for image recognition, we conducted a range of
experiments as described in Table~\ref{tab:Experiments}.
% over batch sizes from 1 to 1024 on benchmark image
% classification data sets. Experiments covered a variety of common model
% architectures (including LeNet \cite{lecun1995learning}, VGG \cite{simonyan2014very}, and ResNet \cite{he2016deep}) run
% on the MNIST \cite{lecun2010mnist}, CIFAR10 \cite{krizhevsky2009learning} and CIFAR100 data sets. The models
% were trained for a fixed number of updates with a slowly decaying
% learning rate. 
Adam \cite{kingma2014adam} adaptive learning rate was also used for one
of the models. Light regularization was used with a decay constant of
$10^{-4}$ on the $L_2$-norm of the weights. For each model
architecture, we varied the size in terms of width (\emph{i.e.}, parameters per
layer) and depth (\emph{i.e.}, number of layers) to measure the training
behavior across model topologies. In addition, we experimented with the
same model across all three data sets (LeNet). Training was performed
using the Torch library on a single K80 GPU.\footnote{None of our
  measurements required data-level parallelism because our decomposition
  of $\TC$ allows us to estimate $\NU$ and $\TU$
  separately, and $\NU$ is independent of
  $P$, the level of data parallelism used.} %Table~\ref{tab:Experiments} summarizes the
%various experiments that were performed. 
Training and cross-validation
losses were recorded after each update for MNIST and after every 100
updates for CIFAR10 and CIFAR100, using two distinct randomly selected
sets of 20\% of the available data. The recorded results were
``scanned'' to find the $\TU$ value
that first achieves the desired training loss level, $\epsilon$. 
This approach is equivalent to a stopping criterion with no patience.

Each MNIST experiment was averaged over ten runs with different random
initializations to get a clean estimate of
$\NU$ as a function of $M$. Averaging was not
used with the other experiments, and as our results show, was not generally
needed.

\begin{figure*}[!htb]
  \centering
  \includegraphics[width=6.10449in,trim={0 11cm 0 0},clip]{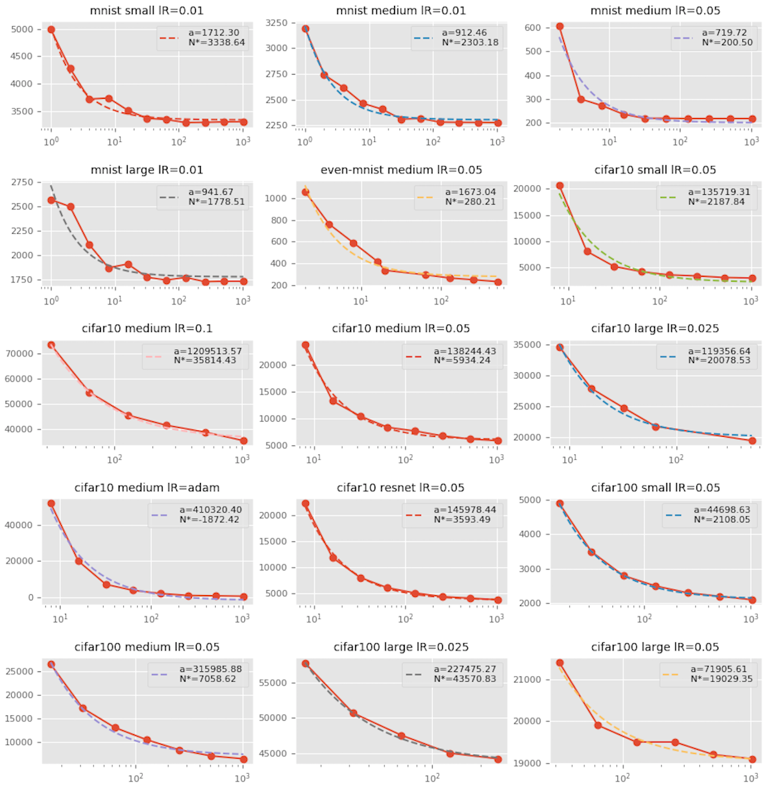}
  \includegraphics[width=6.10449in,trim={0 0 0 5cm},clip]{Figures/ExperimentalResults.png}
  \caption{$\NU$ as a function of $M$ for a
variety of SGD learning problems for a variety of conditions. 
The y-axis represents the training loss. The plots
generally show an inverse relationship between
$\NU$ and $M$. Since $\NU$ is a random variable, we see noise in
these graphs. Additional averaging over multiple runs removes this
noise.}
  \label{fig:ExperimentalResults}
\end{figure*}

The results of our experiments in Figure~\ref{fig:ExperimentalResults} show a robust inverse
relationship between $\NU$ and $M$ measured
across all the data sets, models, learning rates for each case we have
considered. The fit lines match the observed data closely and we
estimated $N_{\infty}$ and $\alpha$. Because of the large number of
possible combinations of experiments performed, we only show a
representative subset of the graphs to illustrate the behavior that was
observed in all experiments. This empirical behavior exists for training error,
cross-validation error, varying $\epsilon$, changing the number of
output classes, etc.

These results show that large learning rates (shown as ``lR'' in the
graphs) are associated with small $N_{\infty}$, which is not
unexpected. However, for the experiment with adaptive learning rate
(cifar10\_medium\_adam), $N_{\infty}$ is negative, which is likely the
result of noise in our estimates or a failure of the model for adaptive
learning rates. Further study is needed to understand this. Even
so, this indicates that $N_{\infty}$ is small compared to the $\alpha$, and
hence good parallel efficiency is possible.

\subsection{Machine Translation Task}
Our translation system implements the attentional model of translation \cite{DBLP:journals/corr/BahdanauCB14} consisting of an encoder-decoder network with an attention mechanism. 
The encoder uses a bidirectional GRU recurrent neural network \cite{DBLP:journals/corr/ChoMBB14} to encode a source sentence ${\bf{x}}=(x_1,...,x_l)$, where $x_i$ is the embedding vector for the $i$th word and $l$ is the sentence length. The encoded form is a sequence of hidden states ${\bf{h}} = (h_1, ..., h_l)$ where each $h_i$ is computed as follows

\begin{equation}
 h_i = 
 \begin{bmatrix}
 \overleftarrow{{h}_i} \\
 \overrightarrow{{h}_i}
\end{bmatrix}
=
\begin{bmatrix}
\overleftarrow{f}(x_i, \overleftarrow{h}_{i+1}) \\
\overrightarrow{f}(x_i, \overrightarrow{h}_{i-1})
\end{bmatrix},
\end{equation}
where $\overrightarrow{h_0} = \overleftarrow{h_0} = 0$.  Here $\overleftarrow{f}$ and $\overrightarrow{f}$ are GRU cells.

Given $\bf{h}$,  the  decoder  predicts  the
target  translation ${\bf y}$  by computing the output token sequence $(y_1,...y_m)$, where
$m$ is the length of the sequence.
At each time $t$, the probability of each token
$y_t$ from a target vocabulary is
\begin{equation}
p(y_t | {\bf h}, y_{t-1}..y_1) = g(s_t, y_{t-1}, H_t),
\end{equation}
where
$g$ is a two layer feed-forward network over the embedding of the
previous target word ($y_{t-1}$), the decoder hidden state ($s_t$), and the weighted sum of encoder states
${\bf h}$ ($H_t$), followed by a softmax to predict the
probability distribution over the output vocabulary.
We use a two layer GRU for $s_t$.
% We compute $s_t$ with a two layer GRU as
% \begin{equation}
% s'_t = r(s_{t-1}, y^*_{t-1})
% \end{equation}
% and
% \begin{equation}
% s_t  = q(s'_t,H_t),
% \end{equation}
% where $s'_t$ is an intermediate state and $s_0=\overleftarrow{h_0}$. 
The two GRU units together with the attention constitute
the conditional GRU layer of \cite{sennrich-EtAl:2017:EACLDemo}.  $H_t$ is computed as
\begin{equation}
H_t =
\begin{bmatrix}
\sum^l_{i=1} (\alpha_{t,i} \cdot \overleftarrow{h}_i) \\
\sum^l_{i=1} (\alpha_{t,i} \cdot \overrightarrow{h}_i)
\end{bmatrix},
\end{equation}
where $\alpha_{t,i}$ are the elements of $\alpha_{t}$ which is the output vector of the attention model.
This is computed with a two layer feed-forward network
\begin{equation}
\alpha'_t = v(\textrm{tanh}(w(h_i) + u(s'_{t-1}))),
\end{equation}
where $w$ and $u$ are weight matrices, and $v$ is another matrix resulting in one real value per encoder state $h_i$.  $\alpha_t$ is then the softmax over $\alpha'_t$.

%\subsection{Machine Translation Experiments}
We train our MT model with the Pytorch framework \cite{paszke2017automatic}.  Learning is
performed with the Pytorch implementation of SGD.  No adjustment to learning
rates was performed. We used the Europarl \cite{Koehn05} German-English data set for training and newstest2013 data set for testing.

\begin{figure}[!htb]
  \centering
  \includegraphics[width = 2.5in]{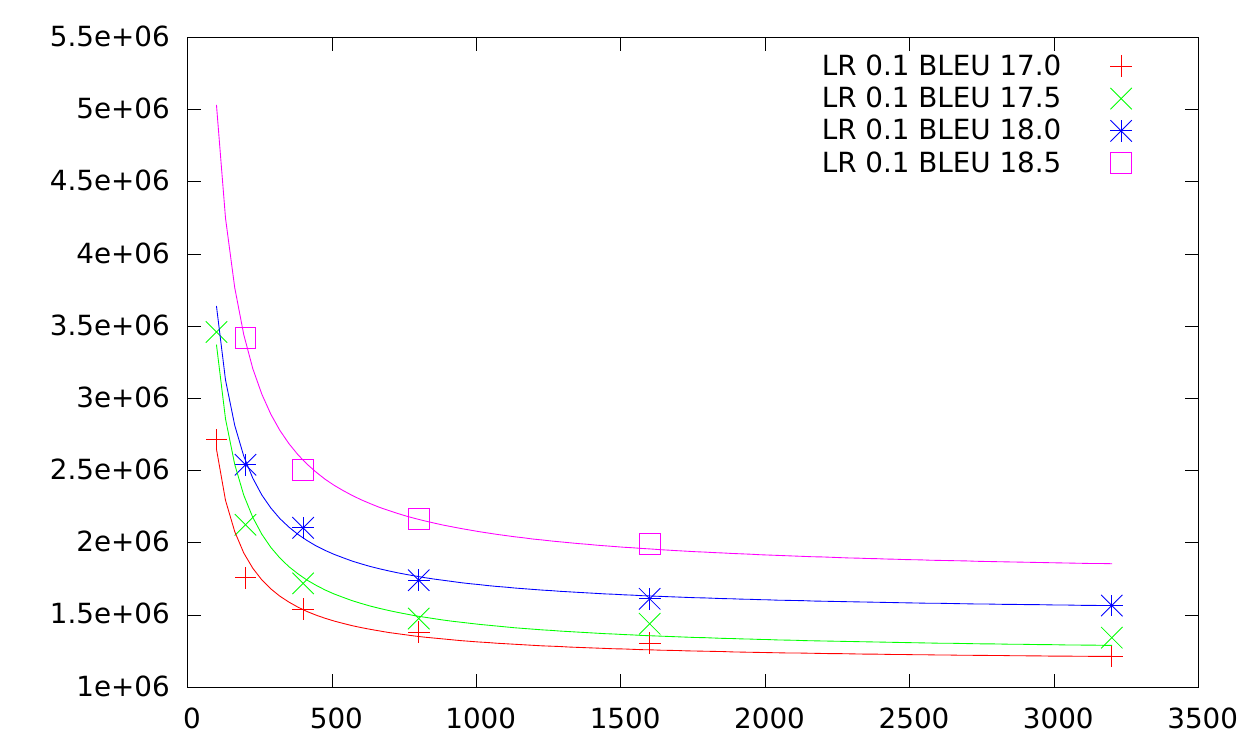}\\
  \includegraphics[width = 2.5in]{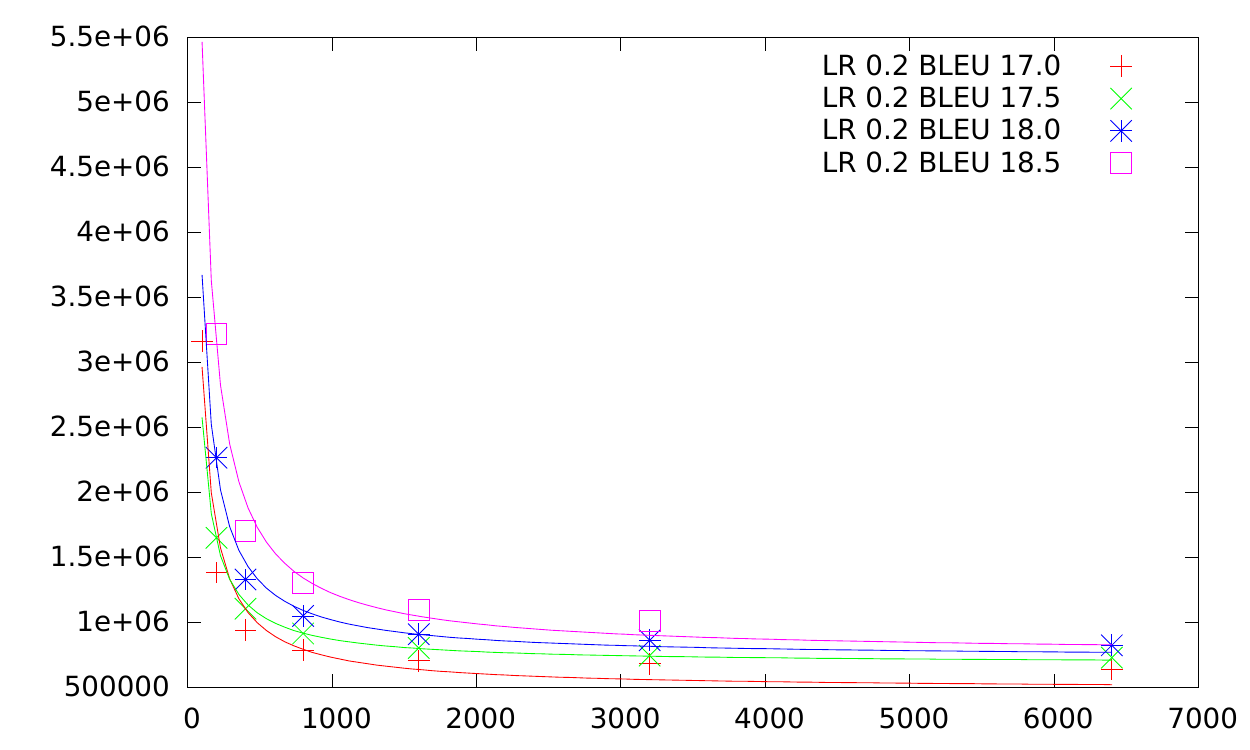}\\
  \includegraphics[width = 2.5in]{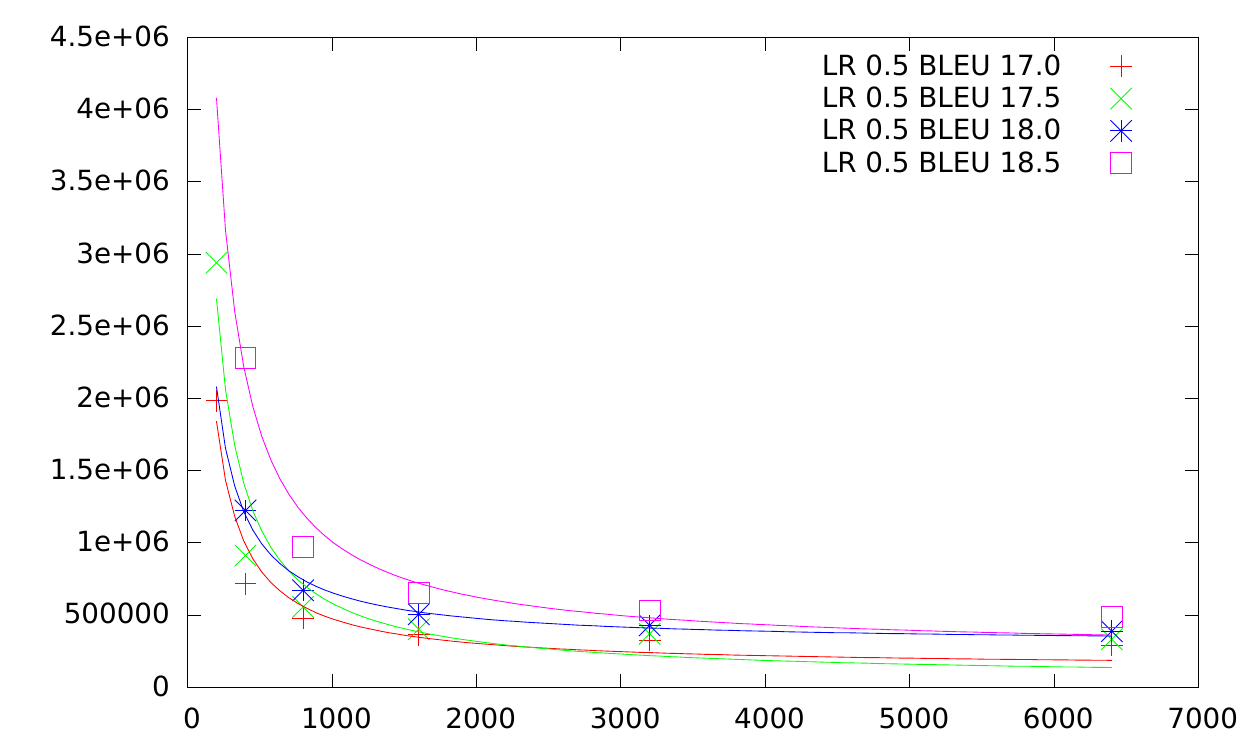}\\
  \includegraphics[width = 2.5in]{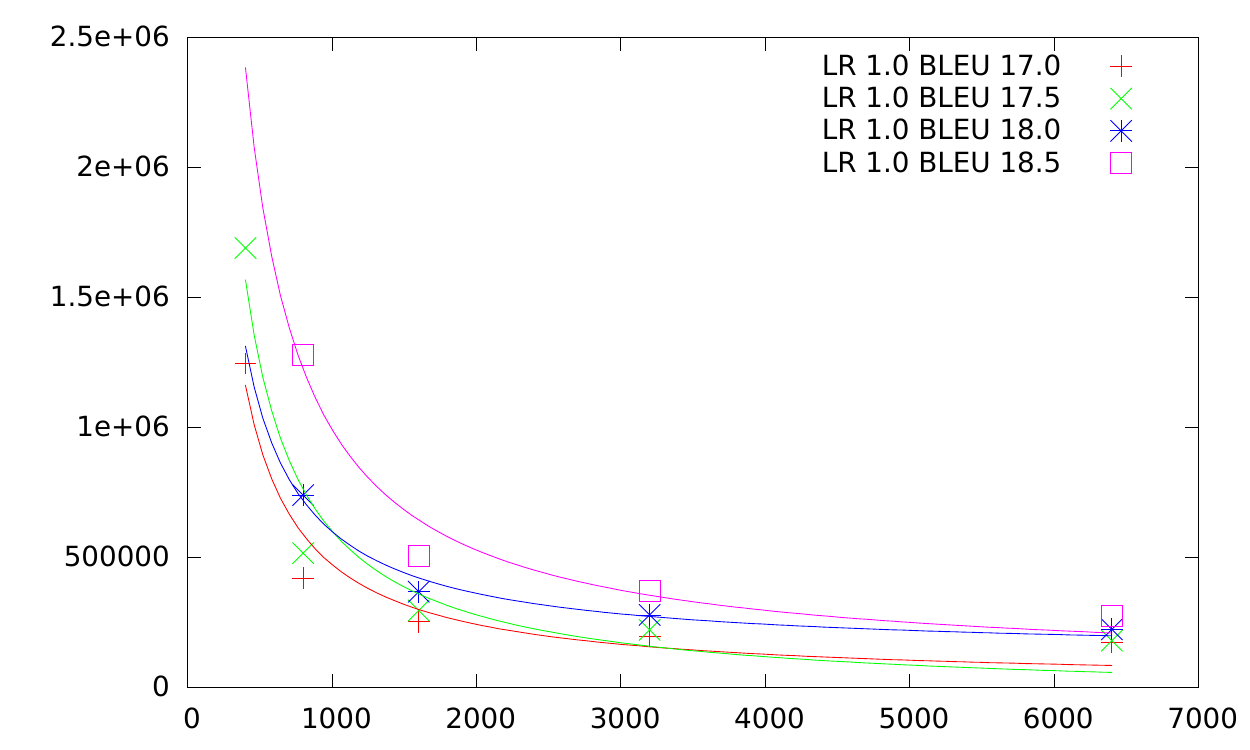}
  \caption{MT results are shown for learning rate, LR, equal to 0.1, 0.2, 0.5, and 1.0.  For four learning error targets (i.e., MT BLEU scores), the measured $N$ values (y-axis) are plotted against their corresponding $M$ values (x-axis).  The solid lines show that this data is well fit by the $\NU=N_\infty+\alpha/M$ model. Points are omitted where training did not achieve a specified BLEU score.}
  \label{fig:MTResults}
\end{figure}

For our experiments, we trained with mini-batch sizes of 100, 200, 400, 800, 1600, 3200, and 6400 words per
mini-batch.  12800 was too large for GPU memory.  Each mini-batch is created by
adding sentences until the number of target language words  meets or exceeds
the batch size.  For each batch size, we ran with learning rates of 0.1, 0.2, 0.5, and 1.0
%2.0, 5.0, and 10.0.  
Each combination of parameters was run with 5 different random
seeds.  The results in Fig.~\ref{fig:MTResults} are again well fit by the $\NU=N_\infty+\alpha/M$ model.

%%%%%%%%%%%%%%%%%%%%%%%%%
%\input{Sections/PracticalConsiderations}

%%%%%%%%%%%%%%%%%%%%%%%%%
\section{Theoretical Bound on $\NU$ for SGD\label{sec:Theory}}

For completion, we provide a theoretical analysis of mini-batch SGD
convergence that supports our finding of a robust empirical inverse relation between $\NU$ and $M$.  We begin by defining the SGD update step as
\begin{equation*}
w^{k + 1} = w^{k} - \eta\left( \nabla f\left( w^{k} \right) + \xi^{k} \right)
\end{equation*}
where $f$ is the function to be optimized, $w^{k}$ is a vector of
neural net weights at the $k^{\rm th}$ update of the SGD algorithm, $\xi^k$
is a zero-mean noise term with variance smaller than $\phi^{2}$, and $\eta$ is
the SGD step size. We assume that $\nabla f$ is Lipschitz continuous,
\emph{i.e.}, that
\begin{equation*}
f\left( x \right) \leq f\left( y \right) + \nabla f\left( y \right) \cdot \left( x - y \right) + \frac{L}{2}\ \left| x - y \right|^{2}
\end{equation*}
for some constant $L$. 
Standard convex analysis steps along with
%Setting $x=w^{k+1}$ and $y=w^{k}$, we get
%\begin{equation}
%f(w^{k+1}) \leq f(w^{k}) + \nabla f(w^{k}) \cdot \left( w^{k+1}-w^{k} \right) + \frac{L}{2}\ \left| w^{k + 1} - w^{k} \right|^{2}.
%\end{equation}
%Inserting the SGD update step in RHS and averaging over $\xi^k$, and using the facts that
$E[\xi^k] = 0$ and Var$[\xi^k] \leq \phi^2$, gives
\begin{small}
\begin{equation*}
E_{\xi^k}\hspace{-0.1cm}\left\lbrack f\hspace{-0.1cm}\left(w^{k+1}\right) \right\rbrack \hspace{-0.1cm}\leq f\left(w^{k}\right) - \eta\left(1-\frac{\eta L}{2} \right)\hspace{-0.1cm}\left| \nabla f\left(w^{k}\right)\right|^{2}\ + \eta^{2}\frac{L}{2}\phi^{2}.
\end{equation*}
\end{small}
We define $\Delta_{k}$ to be the residual at
the $k^{\rm th}$ step, \emph{i.e.},
\begin{equation*}
\Delta_{k} \equiv f\left(w^k \right)-f\left(w^{*}\right)
\end{equation*}
where $w^{*}$ is the global minimum of $f$. 
Using the residual, 
%the
%above inequality becomes
%\begin{equation}
%E_{\xi^{k}}[\Delta_{k+1}] \leq \Delta_{k} - \eta\left(1-\frac{\eta L}{2} \right)\ \left| \nabla f\left( w^{k} \right) \right|^{2}\  + \eta^{2}\frac{L}{2}\phi^{2}.
%\end{equation}
%We assume $f$ is convex, \emph{i.e.},
assuming convexity, we get
%\begin{equation}
%f\left(w^{k} \right) - f\left(w^{*} \right) \leq \nabla f\left( w^{k} \right) \cdot \left(w^{k} - w^{*} \right) \leq \left| \nabla f\left(w^{k} \right) \right| \cdot \left|w^{k} - w^{*} \right|
%\end{equation}
%which implies
\begin{equation*}
\frac{\Delta_{k}}{|w^{0} - w^{*}|} \leq \frac{\Delta_{k}}{\left| w^{k} - w^{*} \right|} \leq \left| \nabla f\left(w^{k} \right) \right|.
\end{equation*}
Choosing the learning rate $\eta$ such that
\begin{equation*}
\left( 1 - \frac{\eta L}{2} \right) > 0
\end{equation*}
results in
\begin{equation*}
E_{\xi^k}[\Delta_{k+1}] \leq \Delta_{k} - \lambda\Delta_{k}^{2} + \lambda\sigma^{2}
\end{equation*}
where
\begin{equation*}
\lambda \equiv \eta\left( 1 - \frac{\eta L}{2} \right)\ \frac{1}{\left( w^{0} - w^{*} \right)^{2}}\text{~~ and~~ }\sigma^{2} \equiv \frac{\eta^{2}L}{2\lambda}\phi^{2}.
\end{equation*}
Now, we take average over the full history and use the fact that
\begin{equation*}
(E[X])^2\leq E[X^2]
\end{equation*}
to obtain
\begin{equation*}
E[\Delta_{k+1}]\leq E[\Delta_{k}]-\lambda(E[\Delta_k])^2+\lambda\sigma^2.
\end{equation*}
For simplicity from here forward, the expectation sign will be omitted by using 
$\Delta_k \equiv E_{\xi^1\xi^2\cdots\xi^{k-1}}[\Delta_k]$. 
We rearrange this inequality as
\begin{equation*}
(\Delta_{k + 1} - \sigma) \leq (\Delta_{k} - \sigma)(1 - \lambda(\Delta_{k} + \sigma))
\end{equation*}
and observing that $\Delta_{k}$ cannot be smaller than $\sigma$
because of constant learning rate and additive noise, implies
\begin{equation*}
1 - \lambda\left(\Delta_{k} + \sigma \right) \geq 0.
\end{equation*}
By taking the inverse and using the fact that
\begin{equation*}
\frac{1}{1 - x} \geq 1 + x \ \ \ \ {\rm for}\ \ \ \ x \leq 1
\end{equation*}
we obtain
\begin{equation*}
\frac{1}{\Delta_{k + 1} - \sigma} \geq \frac{1}{\Delta_{k} - \sigma}\left( \ 1 + \lambda\left( \Delta_{k} + \ \sigma \right) \right) = \frac{1 + 2\lambda\sigma}{\Delta_{k} - \sigma} + \eta.
\end{equation*}
Then, telescoping this recurrence inequality results in
\begin{equation*}
\frac{1}{\Delta_{k + 1} - \sigma} + \frac{1}{2\sigma} \geq \left( 1 + 2\lambda\sigma \right)^{k + 1}\left( \frac{1}{\Delta_{0} - \sigma} + \frac{1}{2\sigma} \right).
\end{equation*}
Finally, solving for $\Delta_{k}$, gives
\begin{equation*}
\label{eqn:PowerDenom}
\Delta_{k} \leq \frac{1}{\left( 1 + 2\lambda\sigma \right)^{k}\left( \frac{1}{\Delta_{0} - \sigma} + \frac{1}{2\sigma} \right)\  - \frac{1}{2\sigma}} + \sigma
\end{equation*}
and, Taylor expanding for small $\sigma$, the number of updates to reach 
$\Delta_{k} \leq \epsilon$ is given by
\begin{align*}
\NU &\geq \frac{\log\left\lbrack \frac{\epsilon + \sigma}{\epsilon - \sigma} \right\rbrack + \log\left\lbrack \frac{\Delta_{0} - \sigma}{\Delta_{0} + \sigma} \right\rbrack}{\log\left\lbrack 1 + 2\lambda\sigma \right\rbrack} \\
&\approx \frac{1}{\lambda}\left( \frac{1}{\epsilon} - \frac{1}{\Delta_{0}} \right)\hspace{-0.1cm}\left( 1 + \frac{\sigma^{2}}{3}\left( \frac{1}{\epsilon^{2}} + \frac{1}{\Delta_{0}^{2}} + \frac{1}{\epsilon\Delta_{0}} \right)\hspace{-0.1cm}\right).
\end{align*}
Using the Central Limit Theorem, we observe that
$\sigma^{2} \approx \frac{\theta}{M}$
and therefore obtain
\begin{equation*}
N_{\text{Update}} \geq \frac{1}{\lambda}\ \left( \frac{1}{\epsilon} - \frac{1}{\Delta_{0}} \right)\hspace{-0.1cm}\left( 1 + \frac{\theta}{M}\left( \frac{1}{\epsilon^{2}} + \frac{1}{\Delta_{0}^{2}} + \frac{1}{\epsilon\Delta_{0}} \right)\hspace{-0.1cm}\right).
\end{equation*}
The fact that this bound exhibits the same inverse $M$ relationship as
\begin{equation*}
\NU = N_{\infty} + \frac{\alpha}{M}
\end{equation*}
reinforces the robustness of our empirical finding.

%%%%%%%%%%%%%%%%%%%%%%%%%
%\input{Sections/ComparisonConvergenceRate}

%%%%%%%%%%%%%%%%%%%%%%%%%
\section{Discussion\label{sec:Summary}}

%The fundamental contribution of this paper is a
%quantitative model of SGD machine learning training time, built on the decomposition of training given by
%\begin{equation}
%\TC=\NU\TU
%\end{equation}
%and the discovery of a robust empirical law relating the number of updates to the mini-batch size:
%\begin{equation}
%\NU = N_{\infty} + \frac{\alpha}{M}.
%\end{equation}
%Our model leads to specific insights and guidance for minimize training time, improving algorithmic development, optimizing system performance and designing new systems. 
%
%Importantly, o
Our model separates algorithmic convergence properties from implementation details. This separation provides machine learning researchers and practitioners a new way of thinking about their algorithms: $N_{\infty}$ provides a lower bound on the number of updates required to converge and fundamentally limits the benefits of parallelization for accelerating machine learning; and $\alpha$ introduces a new concept, that of an algorithm's ``noise sensitivity'', as a key property in the optimization of machine learning.  Using these new principles to guide algorithmic design may help researchers develop improved algorithms.

% Our current research extends our results to new domains.  For example, 
% preliminary results for machine translation using recurrent neural networks are very promising.

% Specifically, this paper described a novel and robust empirical
% relationship between data-parallel scaling behavior and SGD training
% time. This relationship was used to derive optimal scaling for SGD
% machine learning, to define optimal system design, and to provide
% guidance on future algorithmic design.
% \begin{equation}
% \NU = N_{\infty} + \frac{\alpha}{M}
% \end{equation}
% Once the functional forms of $\NU$ and
% $\TU$ are known, the scaling behavior can be
% predicted by minimizing training time over mini-batch size, $M$, for
% a given level of parallelism, $P$. The paper discussed various ways
% of using this equation in this paper using modeling of
% $\NU$ and $\TU$. In
% practice, the $\TU$ can be measured easily; but
% determining $\NU$ requires, in principle, SGD
% iteration until convergence with many different mini-batch sizes, which in general is simply impractical.
% 
% This paper has shown that there exists a robust empirical model of
% $\NU$,
% \begin{equation}
% \NU = N_{\infty} + \frac{\alpha}{M},
% \end{equation}
% which removes this problem. For example, one possibility is to determine
% $\alpha$ and $N_{\infty}$ in the early stage of training and then
% use the fit to choose $M$. Exactly how long to train to get a good
% estimate is not known. More study is needed to explore this and other potential methods for employing this model.

We close with a few observations about challenges and opportunities ahead.

\emph{Noise Sensitivity and Complexity}: Our experiments suggest that as the learning
problem grows in complexity from MNIST to CIFAR10 to CIFAR100, its
sensitivity to noise grows (\emph{i.e.}, $\alpha$ grows). See for example
the medium size model results for fixed learning rate. Thus, the onset
of the $N_{\infty}$ ``floor'' is pushed to larger mini-batch values.
This suggests that the benefit of parallelism may grow as the research
community explores more complex learning challenges. However, this
benefit must be balanced by any related increase in $N_{\infty}$,
which will in general also grow with complexity.

\emph{Improved Learning Algorithms}: The research community should
be encouraged to develop algorithms with lower $N_{\infty}$ as this
will lead to better data-parallel scaling. 
% Algorithms that make better
% use of the data to generate improved update estimates and thereby
% reduce $N_{\infty}$ (\emph{e.g.},perhaps second order methods) are prime
% candidates. Of course, this reduction needs to be understood in the
% context of a tradeoff with a concomitant increase in $\TU$.

% \emph{Local Minima}: It has been shown \cite{keskar2016large} that increasing
% mini-batch size generally has negative effects on generalization.
% Intuitively, the reduced gradient stochasticity of larger mini-batches
% leads to increased risk of getting stuck in local minima. This problem
% is fundamental to data-parallel scaling of SGD and needs to be addressed as
% parallelization efficiency improves.

\emph{Beyond SGD}: %It is important to note that 
The core methodology
presented in this paper is not limited to SGD. 
% It is applicable to any
% algorithm that has a calculation phase followed by a model update phase. 
Research is required to explore whether other robust
mini-batch (or other) relationships exist for different algorithms,
such as Expectation Maximization \cite{dempster1977maximum}. In this way, the methodology
described in this paper provides a new way of comparing the
parallelization effectiveness of algorithms.

%%%%%%%%%%%%%%%%%%%%%%%%%
% \subsection*{Acknowledgements}
% We are grateful to Jerry Quinn for our initial foray into machine translation.

% %%%%%%%%%%%%%%%%%%%%%%%%%
% \section{Additional Considerations}
% \begin{itemize}
% \item
%   \textbf{Hardware Design}: These results indicate that there is no
%   one-size-fits-all for machine learning system design. Each learning
%   problem, model and algorithm will potentially have unique $\alpha$
%   and $N_{\infty}$ values and will benefit from different values of
%   $\delta, \gamma$ and $M_T^2$. Of course, even if a data
%   center has a fixed set of system parameters, one can still optimize
%   the allocation of data center resources based on the methodology
%   presented here.
% \item
%   \textbf{Throughput Parallelization}: This paper has focused on the
%   challenges of parallel training. Once systems are trained, we see no
%   similar fundamental barriers to embarrassingly parallel operation of
%   the trained networks on new data for classification, etc.
% \item
%   \textbf{Enhance Machine Learning Libraries}: Today's machine learning
%   libraries do not provide convenient nor efficient methods for
%   overlapping computation with communication. Developing algorithms and
%   libraries that do so will have significant positive impact on scaling
%   performance.
% \end{itemize}

%%%%%%%%%%%%%%%%%%%%%%%%%
\clearpage
%\begin{small}
\bibliographystyle{apalike}
\bibliography{biblio.bib}

\end{document}

%% file: Sections/DataParallelScaling.tex
%%%%%%%%%%%%%%%%%%%%%%%%%
\section{Data Parallel Scaling of Parallel SGD\label{sec:Scaling}}

Scaling measures the total time to solution, as a function of the number
of computer nodes. 
Traditionally, there are two scaling schemes,
\emph{Strong Scaling} and \emph{Weak Scaling}. We discuss these below
and note that neither is ideal for SGD-based machine learning. We
therefore introduce \emph{Optimal Scaling} and compare the three
approaches.

Our analysis assumes data parallelism, which
% \emph{i.e.}, that the number of data
% samples assigned to each node is an integer. 
% Data parallelism 
leads to
node-level load imbalance (and corresponding inefficiency) when the
minibatch size is not a multiple of the number of nodes, $P$. For
convenience, the analysis below ignores these effects and thus presents
a slightly more optimistic analysis. 
% The alternatives are to take a
% model parallel approach in which a single data sample is split over
% multiple nodes, or a hybrid approach in which both data and model
% parallelism are use. However, model splitting requires additional
% communication and incurs additional computational inefficiencies that
% generally lead to less efficient performance than pure data parallelism.
% This is still an open area of research.

%%%%%%%%%%%%%%%%%%%%%%%%%
\subsection{Strong Scaling}

Strong scaling occurs when the problem size remains fixed. This means
that the amount of compute per node decreases as $P$ increases. For
training tasks, this implies that $M$ is fixed, \emph{i.e.}, $M=M_{\rm Strong}$.
In this case, $\NU$ does not change, so the training time improves only when $\TU$ decreases. Thus, strong scaling hits a
minimum when $P > M_{\rm Strong}/M_T$.

%%%%%%%%%%%%%%%%%%%%%%%%%
\subsection{Weak Scaling}

Weak scaling occurs when the problem size grows proportionately with $P$.
This implies that for training tasks, $M$ grows linearly with $P$
(\emph{i.e.}, $M = mP$) and therefore $\NU$
decreases as $P$ increases, while $\TU$ remains
constant, for constant $m$. Weak scaling can be optimized by
selecting $m$ appropriately, which leads to the optimal scaling
described below.

%%%%%%%%%%%%%%%%%%%%%%%%%
\subsection{Optimal Scaling}
The constant $M$ of strong scaling and the linear $M$ of weak
scaling prevent these methods from achieving optimal performance, and
are therefore inappropriate for SGD-based machine learning. We propose
an alternative approach to scaling that, unlike strong and weak scaling,
minimizes $\TC(M,P)$ over $M$ for each value of $P$. This
approach allows better performance than either strong or weak scaling.

% We can optimize $M$ by considering two cases: $(i)$
% For $M > M_T P$, the optimal $M$ is determined by minimizing
% \begin{equation}
% \TC(M,P) = \left( N_{\infty} + \frac{\alpha}{M}\right)\left(\frac{\gamma M}{P} + \delta \right)
% \end{equation}
% and $(ii)$ for $M \leq M_T P$,
% \begin{equation}
% \TC(M,P) \geq \TC(M_T P,P)
% \end{equation}
% the optimal $M$ is given by $m_T P$. Thus, in general,
% the optimum $M$ is

% \begin{equation}
% M_{\rm Opt}(P) = \max\left( M_T P,\sqrt{\frac{\alpha}{N_\infty}\frac{\delta}{\gamma}P}\right)
% \end{equation}
If we combine Eqn.~\ref{eqn:TC} with Eqn.~\ref{eqn:MOpt}, we get a closed for solution for  the minimum time to convergence:
\begin{equation}
    \TC(P)= 
\begin{cases}
    \left(\sqrt{\strut\delta N_{\infty}} + \sqrt{\strut\frac{\alpha\gamma}{P}} \right)^2, & P < \frac{\alpha\delta}{\gamma M_T^{2}N_{\infty}}\\
    \left(N_{\infty} + \frac{\alpha}{M_{T}P} \right)\left(\delta + \gamma M_T\right), & {\rm otherwise.}
\end{cases}
\end{equation}

Note that for large $P$ (\emph{i.e.}, the second condition above), optimal
scaling is identical to weak scaling. In this
way, optimal scaling naturally defines the per node minibatch size for
weak scaling.

% \begin{figure}[htb]
%   \centering
%   \includegraphics[width=3.43264in,height=2.75903in]{Figures/ComparisonOptWeakStrong.png}
%   \caption{Time to convergence (y-axis on left, solid lines) and
% minibatch size (y-axis on right, dashed lines) as a function of $P$ for
% Strong (red), Weak (blue) and Optimal (green) Scaling. Optimal
% Scaling is always the best, but that when $P$ is large enough, Weak and
% Optimal are equivalent.}
%   \label{fig:ScalingComparison}
% \end{figure}

%% file: Sections/SystemDesign.tex
%%%%%%%%%%%%%%%%%%%%%%%%%
\subsection{Optimal System Hardware Design\label{sec:SystemDesign}}
We now show how optimal scaling can be used to optimize learning system hardware design.
The principle behind optimal system design is to balance the trade-offs between various system parameters so as to optimize some system performance metric, like time to convergence.  If we assign a cost for each system component, such as the number of nodes, the size of the communication network, the amount of memory, etc., we can then find the value of these elements that optimize the system metric.  This approach can be used to optimize system design for multiple machine learning problems, e.g., determining the correct ratio of compute to communication in a system; or to efficiently allocate resources in a data center running multiple learning problem concurrently, e.g., to decide how many learners to apply to each learning task running in a data center.

To make this explicit, consider optimizing a system for bandwidth and compute only.  If we have a fixed amount of money to spend, then the cost, $C$, for each component must be constrained by
\begin{equation}
C_{\rm Compute}(P,\gamma) + C_{\rm Bandwidth}(\delta) = {\rm constant.}
\end{equation}
Combining this constraint with $\TC$ implies that optimal design occurs when the mix of compute and bandwidth satisfy
\begin{equation}
\frac{{\Delta}\TC(M_{\rm Opt}(P),P,\gamma)}{{\Delta}C_{\rm Compute}(P,\gamma)} = \frac{{\Delta}\TC(M_{\rm Opt}(P),P,\gamma)}{{\Delta}C_{\rm Bandwidth}(\delta)}.
\end{equation}
In other words, performance gain per unit price must be balanced at the
optimal design point. Since we have a closed form solution for $\TC$, we can easily find this optimal point if given the various costs.  This approach can be generalized to optimize of multiple constraints.